\documentclass[11pt]{article}
\usepackage{mtsummit2019}
\usepackage{times}
\usepackage{url}
\usepackage{latexsym}
\usepackage[small,bf]{caption} 
\usepackage{multirow}
\usepackage{algorithm}
\usepackage{algpseudocode}
\usepackage{amsmath}
\usepackage{xcolor}
\usepackage[colorinlistoftodos,prependcaption,textsize=tiny]{todonotes}
\usepackage{comment}
\definecolor{brgreen}{rgb}{0.0, 0.26, 0.15}
\usepackage{natbib}
\usepackage{enumitem}

\title{Improving CAT Tools in the Translation Workflow:\\ New Approaches and Evaluation}

\author{Mihaela Vela\textsuperscript{1}, Santanu Pal\textsuperscript{1,2}, Marcos Zampieri\textsuperscript{3} Sudip Kumar Naskar\textsuperscript{4} \\ \bf{Josef van Genabith\textsuperscript{1,2}} \vspace{2mm} \\ 
  \textsuperscript{1}Saarland University, Germany, \textsuperscript{2}DFKI, Germany\\
  \textsuperscript{3}University of Wolverhampton, UK, \textsuperscript{4}Jadavpur University, India \\
   {\tt m.vela@mx.uni-saarland.de}}

\date{}

\begin{document}
\maketitle
\begin{abstract}
This paper describes strategies to improve an existing web-based computer-aided translation (CAT) tool entitled {\em CATaLog Online}. 
{\em CATaLog Online} provides a post-editing environment with simple yet helpful project management tools. 
It offers translation suggestions from translation memories (TM), machine translation (MT), and automatic post-editing (APE) and records detailed logs of post-editing activities. 
To test the new approaches proposed in this paper, we carried out a user study on an English--German translation task using {\em CATaLog Online}. 
User feedback revealed that the users preferred using {\em CATaLog Online} over existing CAT tools in some respects, especially by selecting the output of the MT system and taking advantage of the color scheme for TM suggestions.

\end{abstract}

\section{Introduction}

The use of computer software is an important part of the modern translation workflow~\citep{zaretskaya2015,schneidersurvey2019}.
A number of tools are widely used by professional translators, most notably CAT tools and terminology management software. These tools increase translators' productivity, improve consistency in translation and, in turn, reduce the cost of translation~\citep{zampieri2014}.
The most important component in state-of-the-art CAT tools are translation memories (TM). 
The translators can either accept, reject or modify the suggestions received from the TM engine. As the process is done iteratively, every new translation increases the size 
of the translation memory making it more useful for future translations. 

The idea behind TMs is relatively simple, however, the process of matching and retrieval of source and target segments is not trivial. 
In this paper we discuss new approaches to improve TM retrieval and CAT tools interfaces. 
With our contribution we aim to make TM suggestions more useful and accurate
\begin{enumerate}[label=(\roman*)]
\item by presenting new retrieval strategies for the TM suggestions, and
\item by making the translator's job easier in terms of presenting the translation suggestions in the CAT tool.
\end{enumerate}

\noindent To achieve these goals, we use a new web-based CAT tool called {\em CATaLog Online} \citep{Pal:2016}\footnote{Available at~\url{http://santanu.appling.uni-saarland.de/MMCAT/}}, 
which builds on an existing desktop CAT {\em CATaLog} \citep{Nayek:2015} 
but is enhanced with with a new interface layout.

The remainder of this paper is structured as follows: Section \ref{sec:rw} presents related work on CAT tools and TMs, Section~\ref{sec:tool} describes the main functions of {\em CATaLog Online} including similarity matching, color coding scheme, and strategies to improve TM search efficiency. 
Section~\ref{sec:user} presents the results obtained in the user studies carried out, and finally Section~\ref{sec:conclusion} presents the conclusions of this paper and avenues for future research.
\section{Related Work}
\label{sec:rw}
Most professional translators today use the so-called computer-aided translation (CAT) tools~\citep{van2015recommendations,schneidersurvey2019}.
General-purpose CAT tools offer a variety of features, most commonly TM, MT, a glossary and terminology management, concordance search to display words in context, quality estimation (QE) check, QE scores, auto-completion suggestions, and several administrative features to organize projects.

In the translation and localization industry, translators are more and more acting as post-editors, working with pre-translated texts from TM or MT output. This has turned CAT tools an essential part of the translators' workflow. A number of studies on translation process were carried out to investigate translators' productivity, cognitive load (CL), effort, time, quality, etc.

\cite{guerberof12} and~\cite{zampieri2014} report on studies comparing the productivity and quality of human translations using MT and TM output, showing the gain in productivity when post-editing MT segments in comparison to using TM segments or when translating from scratch. 
The incorporation of MT output into the CAT tools allows also for a different kind of MT evaluation.
\cite{Anna:2016b,Anna:2016a} approached post-editing and MT output from a different perspective, namely by using post-editing indicators and the post-editing environment (a CAT tool) to reason about the difficulty of MT output. 
In her overview on the existing methods for measuring post-editing effort (identified by temporal, technical and cognitive indicators)~\cite{koponen2016}, concluded that determining the amount of cognitive effort still poses questions. She further argued that accurate measurements would influence productivity, but the individual experience of the post-editors as well as their work conditions are also criteria to be considered.

TM as a feature is still valued higher than MT, with 75\% of translators believing it to increase throughput and preserve consistency, while 40\% think MT usage is problematic due to the amount of errors~\citep{moorkens2017assessing}.
The retrieval of TM matches in most commercial and many research systems are based on string matching mechanisms that do not exploit semantic similarity~\citep{gupta2015,Gupta2016} and post-editing effort~\citep{koponen2012comparing}, and the presentation of TM matches to users touches upon a research topic in human–computer interaction (HCI) – information visualisation – that has received little attention in both translation studies (TS) and natural language processing (NLP).
\cite{obrien2012} views translation as a form of human-computer interaction showing how the translation profession has changed over time, also due to the newest developments in the area of machine translation and the integration of the MT output into CAT tools for post-editing.
This view is mirrored in recent research, dealing with cognitive load in the translation and post-editing process.
\cite{vieira2014} uses a psychology-motivated definition of cognitive load, while~\cite{Herbig2019} propose a model that uses a wide range of physiological and behavioral sensor data to estimate perceived cognitive load  during post-editing of machine MT.

These findings suggest that a) MT is definitively suitable to be integrated into a TM, b) even a slightly better MT output integrated into a translation environment can improve the translation performance and c) post-editing indicators should consider - if possible - also the personal performance of each translator.




\section{{\em CATaLog Online}: System Description}
\label{sec:tool}
This section describes the {\em CATaLog Online}, a novel and user-friendly web-based CAT tool, its main functionalities and novel features that distinguish it from other CAT tools.
{\em CATaLog Online} offers translations from three engines -- TM~\citep{Nayek:2015}, MT~\citep{Pal:2015:WMT-T} and APE~\citep{Pal:2015:WMT}, from which users can choose the most suitable translation and post-edit.
Users can upload their own translation memories to the platform or can make use of the background translation memory, if any, integrated into the tool for the language pair.
Instead of using the background MT tools, users can also upload the translations produced by third-party MT systems.

\paragraph{TM Search and Segment Retrieval}
\label{similarity}

{\em CATaLog Online} combines elements of both TER and Needleman-Wunsch algorithm to design its similarity and retrieval metric. We take the alignment computed by TER but calculate the similarity score using the intuition of the Needleman-Wunsch algorithm by penalizing edit operations and rewarding matches. A detailed description of TM retrieval implemented in {\em CATaLog Online} can be found in~\cite{Nayek:2015}.

\paragraph{Color Coding}
\label{coloring}

To make that decision process easy, {\em CATaLog Online} color codes the matched and unmatched parts in both source and target sides of the TM suggestions. Green portions imply that they are matched fragments and red portions imply mismatches. 

Ideally, the TM suggestion translation having the maximum number of green words should be the ideal candidate for post-editing.

\paragraph{Improving Search Efficiency} 
\label{search}


Comparing every input sentence against all the TM source segments makes the search process very slow, particularly for large TMs. To improve search efficiency, {\em CATaLog Online} uses the Nutch\footnote{\url{http://nutch.apache.org/}} information retrieval (IR) system. 
Nutch follows the standard IR model of Lucene\footnote{\url{http://lucene.apache.org/}} with document parsing, document Indexing, TF-IDF calculation, query parsing and finally searching/document retrieval and document ranking. In our implementation, each document contains (a) a TM source segment, (b) its corresponding translation and (c) the word alignments.

\paragraph{Machine Translation and Automatic Post Editing}
\label{mt}
Along with TM matches, {\em CATaLog Online} provides MT output~\citep{Pal:2015:WMT-T} to the translator, an option provided by many state-of-the-art CAT tools (e.g. MateCat~\citep{matecat}). Besides the retrieved TM segment and the MT output {\em CATaLog Online} provides also a third option to the translator: the output of an automatic post-editing system meant to be post-edited as the MT output.
The APE system is based in an OSM model~\citep{pal2016usaar} and proved to deliver competitive performance in previous editions of the Automatic Post Editing (APE) shared task at WMT \cite{bojar-EtAl:2016:WMT1}.

\paragraph{Editing Logs}
For a given input segment, {\em CATaLog Online} provides four different options: TM, MT, APE and translation from scratch; the translator either chooses the best translation suggestion among these options or translates from the scratch. For both post-editing and translation the CAT tool the user activities are logged and can be downloaded in XML format. In addition to these logs, the translator can also download the alignments between source and target text.



\paragraph{Data}
\label{sec:data}
The data used for building the internal TM in {\em CATaLog Online} as well as MT and APE system consists of the EuroParl corpus and the news and common crawl corpus collected during the 2015 WMT shared. task\footnote{\url{http://www.statmt.org/}}
\section{User Studies with {\em CATaLog Online}}
\label{sec:user}

We conducted experiments with Translation Studies students and professional translators to evaluate {\em CATaLog Online}. The data used in the evaluation process was translated from English into German.
The goals of our user studies are:
\begin{enumerate}[label=(\roman*)]
\item to compare {\em CATaLog Online} and a similar CAT tool, MateCAT, in terms of human post-editing performance;
\item to compare the efficiency of the three proposed solutions (TM, MT and APE) in a real translation environment.
\end{enumerate}

\noindent The comparison between MateCat and {\em CATaLog Online} was carried out by students performing post-editing on English to German MT output. The 16 students participating in this evaluation were undergraduate students enrolled on a Translation Studies program, attending a translation technologies class, including sessions on MT and MT evaluation.
All of them were native speakers of German, with no professional experience, but with good or very good knowledge of English (B2 and C1 level\footnote{Linguistic competence categories as in the Common European Framework: \url{https://www.coe.int/en/web/common-european-framework-reference-languages/level-descriptions}}). 

Half of the students were asked to perform post-editing of the MT output in MateCat, the other half in {\em CATaLog Online}.
Each student was presented with a set of 30 sentences (news in English and the corresponding German MT output) and was asked to perform post-editing on the German MT output.
From the set of 30 sentences, 20 sentences were randomly chosen, 10 sentences were common to all students, allowing the direct comparison between MateCat and {\em CATaLog Online}.

MateCat captures information about the number of words, the post-editing time and effort, but is also tracking the changes between the MT output and the final post-edited version of the MT output.
{\em CATaLog Online} captures information about post-editing time, and also keeps track of the changes, counting the number of insertions, deletions, substitutions, and shifts.


Since post-editing time (measured in seconds) is the information captured by both tools, we are using it for the comparison between Matecat and {\em CATaLog Online}.
This contrasting listing of the post-editing times holds just for the 10 sentences in common, where we can be sure that the sentences have the same length.

Table~\ref{tab:compare10} shows the post-editing time in seconds, proving that the sentences in MateCat were edited faster than in {\em CATaLog Online}.
The notation S1 to S16 stands for each of the 16 evaluators.
One reason for this result, also commented by the evaluators, might be the different design of the editing interface.
MateCat provides a plain, simple interface, whereas {\em CATaLog Online}'s interface is quite colorful containing more than just editing window.

\begin{table}[ht]
\centering
\small
	\begin{tabular}{l|c||l|c}
    	\hline
        &	{\bf MateCat}	& &	{\bf {\em CATaLog Online}}	\\
  
		\hline
		Stud1	&	1112	&	Stud9		&	3079	\\
		Stud2	&	1086	&	Stud10		&	2623	\\	
        Stud3	&	1304	&	Stud11		&	1761		\\
		Stud4	&	2602	&	Stud12		&	5499	\\	
       	Stud5	&	2176	&	Stud13		&	1788		\\
		Stud6	&	876		&	Stud14		&	5773	\\	
        Stud7	&	901		&	Stud15		&	3040		\\
		Stud8	&	823		&	Stud16		&	4178	\\	
		\hline
	\end{tabular}
	\caption{Direct comparison of MateCat and {\em CATaLog Online} by post-editing time (in seconds) for the 10 sentences in common.}
	\label{tab:compare10}
\end{table}

\noindent The second experiment is addressing the quality of the proposed translation solutions in {\em CATaLog Online}. 
Users are provided with the following translations:
\begin{itemize}
\setlength\itemsep{0.5em}
\item[$\bullet$] the translation from {\em CATaLog Online}'s TM,
\item[$\bullet$] the output of the integrated machine translation system,
\item[$\bullet$] the output of the integrated automatic post-editing system
\end{itemize}

\noindent In order to evaluate the three proposed solutions (TM, MT and APE) in a real translation environment, the same 16 students from the post-editing task were asked to select the most helpful translation.
The experimental design was similar to the one above.
Each student was presented 30 English news sentences in {\em CATaLog Online}, 10 being in common to all students, and asked to opt for the most appropriate German translation.
In the evaluation phase of this experiment, we noticed that the students' decision for the MT or APE system is based on chance, since the MT output and the output from the APE system are very similar to each other.
As a consequence, we excluded the APE output from the list of possible translations and repeated the experiment with three professional translators.
The professional translators were native speaker of German with at least two years of experience in translation. 
Before translating they were provided with guidelines and a short introduction into working with {\em CATaLog Online}.
The translators were asked to perform English to German translation of 200 news sentences with {\em CATaLog Online} by choosing between:

\begin{enumerate}[label=(\alph*)]
\item {the output of {\em CATaLog Online}'s MT system (MT), }
\item {the suggestions from {\em CATaLog Online}'s internal translation memory (TM), }
\item {translating from scratch without any suggestion (None).}
\end{enumerate}

\noindent The selection of the first two possibilities (a) or (b) assumes that translators will edit suggestions proposed by the tool, while for (c) he/she will have to do the translation from scratch.  

\begin{table}[!ht]
\centering
\small
	\begin{tabular}{cccc||ccc}
		\hline
        &	\multicolumn{3}{c||}{\bf 200 sentences}	& 	\multicolumn{3}{c}{\bf 100 sentences}		\\
		& T1 & T2 & T3 &   T1 & T2	& T3	\\
		\hline
		MT		&	160		&	169	&	161	&	74		&	85	&	82\\
		TM		&	1		&	16	&	0	&	1		&	7	&	0\\	
		None	&	39		&	15	&	39	&	25		&	8	&	18\\     
		\hline
	\end{tabular}
	\caption{Selection of suggestions by translators in {\em CATaLog Online}.}
	\label{tab:engineAll}
\end{table}

\noindent From the set of 200 sentences each translator received, 100 were repeated, allowing us to measure the agreement between the three translators.
Since {\em CATaLog Online} is providing an extensive editing log, we collected  information concerning the engine used in translation (MT, TM, or translation from scratch), the number of deletions, insertions, substitutions and shifts as well the edit time (in seconds) for each segment.

\begin{table*}
\centering
\small
	\begin{tabular}{cccc||ccc||ccc}
		\hline
         &	\multicolumn{3}{c||}{\bf Selected suggestions}	& 	\multicolumn{3}{c||}{\bf Editing time}	& 	\multicolumn{3}{c}{\bf Number of edits} \\
 
     		& T1 & T2 & T3 & T1	& T2 & T3 & T1 & T2	& T3 \\
		\hline
		T1&-&0.08&0.20&-&-0.16&-0.06&-&0.49&0.42\\
		T2&0.08&-&0.05&-0.16&-&-0.13&0.49&-&0.26\\
		T3&0.20&0.05&-&-0.06&-0.13&-&0.42&0.26&-\\
		\hline
	\end{tabular}
	\caption{Cohen's $\kappa$ measuring agreement for the selected suggestion, editing time and number of edits.}
	\label{tab:agree}
\end{table*}

The first analysis of the logs shows that all three translators have a tendency in choosing first the suggestion made by the MT system and perform further editing on it. 
Table~\ref{tab:engineAll} gives an overview of the selected suggestions and shows that the MT system achieves a selection rate of around 80\%. 
The remaining sentences are either translated from scratch or by using the suggestions provided by the TM.
The selection suggestions are similar for both the 200 sentenced and the 100 sentences in common.

For the 100 sentences in common, we measured pairwise inter-rater agreement between translators by computing Cohen's $\kappa$~\cite{Cohen:1960} for different variables.
We concentrated on the suggestions used in the translation process (MT, TM, or translation from scratch), editing time, as well as the overall number of edits.

From Table~\ref{tab:agree}, we observe that translators agree only in terms of overall number of edits.
Editing time and the selection of a specific suggestion (MT, TM, or translation from scratch) are parameters on which the translators do not agree.
We computed Pearson's correlation coefficient $\rho$, to test whether the total number of edits (with a low $\kappa$) is influencing the post-editing time (with a high $\kappa$).
We achieved a $\rho$ value of 0.10, not allowing us for a clear interpretation concerning correlation. 

\begin{figure*}[!ht]
\begin{center}
\includegraphics[scale=0.75]{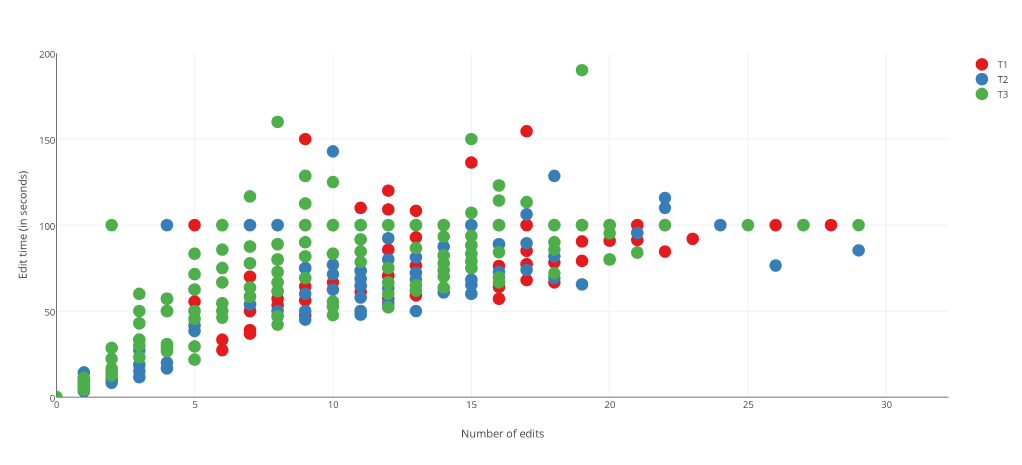}
\caption{Correlation between the overall number of edits and edit time.}
\label{fig:CorrEditTimeVsEdits}
\end{center}
\end{figure*}

Figure~\ref{fig:CorrEditTimeVsEdits}, depicts a slight tendency that a higher number of edits requires more edit time.
We also notice cases in which a high number of edits do not require much editing time and vice versa.
It seems that a higher number of edits does not necessary mean a longer editing time, this being an indicator for the fact that post-editing time is a subjective measure and should be treated carefully. 

Taking a closer look at the type of edits performed during editing, 
we notice that the edits with the highest frequency are substitutions, followed by insertions, deletions and shifts. 
Concluding on the user studies described in this section, we show that translators have a clear preference in choosing the output of the MT system for performing their translation task, even if they do not make the same decision for the same segments.
In terms of editing time, the data shows that in this setting, time is a translator-dependent variable, influencing the low correlation coefficient with the number of edits.
This aspect has to be taken into consideration when measuring post-editing/translation effort by post-editing/translation time, since time is a subjective measure for effort depending on the experience level, working conditions as well as personal abilities.

\subsection{User Feedback}
The professional translators participating in our experiment were asked to rate {\em CATaLog Online} by comparing it to other CAT tools in terms of usability.
The main positive and negative impressions are summarized below.
\paragraph{Positive Feedback}
Translators reported that the unique coloring system in {\em CATaLog Online} - offered by none of the existing TM based CAT tools - helped them to complete the editing of suggestions from the TM. 
They also found the proposed MT suggestions as really helpful and referred positively to the arrangement of the suggestions in {\em CATaLog Online}.


\paragraph{Negative Feedback}
The lack of certain functionalities like a spell-checker, keyboard shortcuts, a concordancer was rated negatively by the translators. 
Although they rated positively the color coding, the interface was considered to be overloaded.   
\subsection{Limitations}
\label{limitation}

Finally, based on the experiments carried out and the feedback from participants we believe that the current version of {\em CATaLog Online} has the following limitations:
\begin{itemize}
    \item Currently, the tool cannot handle document formatting such as bold/italic fonts, bullets;
    \item It does not handle stemming;
    \item The current experiment does not consider individual edit operations in terms of coherence and cohesion of the whole segment which calls for a controlled experiment towards this specific objective by defining different test set for each individual edit operations. 
\end{itemize}

\section{Conclusions and Future Work}
\label{sec:conclusion}
The paper presents strategies to improve a new free open-source  CAT tool and post-editing interface, {\em CATaLog Online}, based on several experiments carried out and presented in this paper. The tool offers translation suggestions from TM, MT and APE.
The tool is specifically designed to improve post-editing productivity and user experience with CAT. A novel feature in the tool is a new intra-segment color coding scheme that highlights matching and irrelevant fragments in suggested TM segments. 
The feedback from the translators show that color coding the TM suggestions makes the decision process easier for the user as to which TM suggestion to choose and work on. 
It also guides the translators as to which fragments to post-edit on the chosen TM translation.
The similarity metric employed in the tool makes use of TER, Needleman–Wunsch algorithm and Lucene retrieval score to identify and re-rank relevant TM .
The tool keeps track of all the post-editing activities and records detailed logs in well structured XML format which is beneficial for incremental MT/APE and translation process research. 
The {\em CATaLog Online} user evaluation
showed that translators have a clear preference in choosing the output of the MT system for performing their translation task. They also evaluated positively the color scheme for the TM suggestions as well as the arrangement of the suggestions within the tool. The informal feedback revealed that features like spell-checker, quality assessment (QA) features and keyboard shortcuts could improve the tool further.


\section*{Acknowledgments}

We would like to thank the participants of this user study for their valuable contribution. We further thank the MT Summit anonymous reviewers for their insightful feedback. 

This research was funded in part by the German research foundation (DFG) under grant number GE 2819/2-1 (project MMPE) and People Programme (Marie Curie Actions) of the European Union's Framework Programme (FP7/2007-2013) under REA grant agreement no 317471. We are also thankful to Pangeanic, Valencia, Spain for kindly providing us with professional translators for these experiments.

\bibliographystyle{apalike}
\bibliography{tmbib.bib}

\end{document}